\documentclass[conference]{IEEEtran}
\IEEEoverridecommandlockouts
\usepackage{cite}
\usepackage{amsmath,amssymb,amsfonts}
\usepackage{algorithmic}
\usepackage{graphicx}
\usepackage{textcomp}
\usepackage{xcolor}
\usepackage{todonotes}
\usepackage{hyperref}
\usepackage{xcolor}
\usepackage{hyperref}
\hypersetup{
    colorlinks,
    linkcolor={red!50!black},
    citecolor={blue!50!black},
    urlcolor={blue!80!black}
}

\usepackage{soul}
\usepackage{arabtex} 
\usepackage{utf8}
\setcode{utf8}

\usepackage{booktabs}
\usepackage{multirow}
\usepackage{float}

\usepackage{hyperref}

\usepackage{booktabs} 

\usepackage{tikz}
\newcommand{\tikzxmark}{%
\color{black}
\tikz[scale=0.1] {
    \draw[line width=0.7,line cap=round] (0,0) to [bend left=6] (1,1);
    \draw[line width=0.7,line cap=round] (0.2,0.95) to [bend right=3] (0.8,0.05);
}}
\newcommand{\tikzcmark}{%
\color{black}
\tikz[scale=0.15] {
    \draw[line width=0.7,line cap=round] (0.25,0) to [bend left=10] (1,1);
    \draw[line width=0.8,line cap=round] (0,0.35) to [bend right=1] (0.23,0);
}}

\usepackage{float}

\lineskiplimit=-\maxdimen

\usepackage{textcomp} 

\def\BibTeX{{\rm B\kern-.05em{\sc i\kern-.025em b}\kern-.08em
    T\kern-.1667em\lower.7ex\hbox{E}\kern-.125emX}}

\usepackage{tabularx}
\newcolumntype{Y}{>{\centering\arraybackslash}X}
\newcolumntype{R}{>{\right\arraybackslash}X}

\begin{document}
\title{Open Automatic Speech Recognition Models for Classical and Modern Standard Arabic}
\author{
\IEEEauthorblockN{Lilit Grigoryan}
\IEEEauthorblockA{\textit{NVIDIA} \\
Yerevan, Armenia \\
lgrigoryan@nvidia.com}
\and
\IEEEauthorblockN{Nikolay Karpov}
\IEEEauthorblockA{\textit{NVIDIA} \\
Yerevan, Armenia \\
nkarpov@nvidia.com}
\and
\IEEEauthorblockN{Enas Albasiri}
\IEEEauthorblockA{\textit{NVIDIA} \\
Santa Clara, USA \\
ealbasiri@nvidia.com}
\and
\IEEEauthorblockN{Vitaly Lavrukhin}
\IEEEauthorblockA{\textit{NVIDIA} \\
Santa Clara, USA \\
vlavrukhin@nvidia.com}
\and
\IEEEauthorblockN{Boris Ginsburg}
\IEEEauthorblockA{\textit{NVIDIA} \\
Santa Clara, USA \\
}
}

\maketitle

\begin{abstract}
Despite Arabic being one of the most widely spoken languages, the development of Arabic Automatic Speech Recognition (ASR) systems faces significant challenges due to the language's complexity, and only a limited number of public Arabic ASR models exist. While much of the focus has been on Modern Standard Arabic (MSA), there is considerably less attention given to the variations within the language. This paper introduces a universal methodology for Arabic speech and text processing designed to address unique challenges of the language. Using this methodology, we train two novel models based on the FastConformer architecture: one designed specifically for MSA and the other, the first unified public model for both MSA and Classical Arabic (CA). The MSA model sets a new benchmark with state-of-the-art (SOTA) performance on related datasets, while the unified model achieves SOTA accuracy with diacritics for CA while maintaining strong performance for MSA. To promote reproducibility, we open-source the models and their training recipes.

\end{abstract}


\section{Introduction}
Although Automatic Speech Recognition (ASR) has advanced significantly for languages like English and Mandarin, Arabic remains particularly challenging due to its unique linguistic features. Moreover, while there are several open Arabic ASR models available, their performance is generally limited, highlighting the need for continued advancement in this field. 

Arabic script mainly consists of consonants and long vowels, with short vowels frequently omitted in written text. Diacritics, representing these short vowels, provide essential phonetic guidance for accurate pronunciation and meaning. Although native speakers can infer the correct meaning from context, the absence of diacritics increases ambiguity, particularly for non-native speakers and language learners.

Arabic has three linguistic variations, each with distinct features. Modern Standard Arabic (MSA) is the standardized written form of the language used in formal communications, educational materials, and news. Diacritics in MSA texts are often omitted, except in cases where they are needed to clarify ambiguity. However, Dialectal Arabic (DA) is used in daily spoken communications, which varies widely and lacks standardized orthography. Classical Arabic (CA) is the language used in the Quran and Islamic literature. CA text features a more complex structure compared to MSA and is usually fully vowelized, meaning that it includes all diacritical marks. 

Developing a robust Arabic ASR system requires researchers to have not only technical expertise but also an understanding of Arabic linguistic features. In this paper, we present two new open Arabic ASR models and a detailed exploration of building pipelines, addressing the unique challenges of the language. Our main contributions are:
\begin{itemize}
    \item Description of two Arabic language variations (MSA and CA) and a thorough analysis of the available open datasets.
    \item Development of a universal pipeline for Arabic speech data preprocessing, made available as open-source.\footnote{
\url{https://github.com/NVIDIA/NeMo-speech-data-processor/tree/main/dataset_configs/arabic}}
    \item Two SOTA Arabic ASR models\footnote{
\url{https://hf.co/nvidia/stt_ar_fastconformer_hybrid_large_pc_v1.0}}\footnote{
\url{https://hf.co/nvidia/stt_ar_fastconformer_hybrid_large_pcd_v1.0}} for both MSA and CA, available under a commercial license to support ongoing research and practical applications.
\end{itemize}

Our models were developed using the open-source NeMo framework \cite{toolkit:nemo} for Conversational AI. The model checkpoints are released on HuggingFace.


\section{Related Work}

\subsection{Arabic ASR}
Early works on Arabic ASR employed traditional ASR systems such as Hidden Markov Models (HMM) \cite{article:hmm1, article:hmm2, article:hmm3, article:toolkit_kaldi_recipe}, Gaussian Mixture Models (GMM) \cite{article:gmm1}, Deep Neural Networks (DNN) \cite{article:dnn1, article:dnn2, article:toolkit_kaldi_recipe} and their hybrids \cite{article:hybrid1, article:hybrid2, article:hybrid3} for Acoustic modeling. More recently, end-to-end models have been adopted for this task \cite{article:e2e1, article:e2e3, article:e2e4, article:e2e5, article:e2e6, article:e2e7, article:e2e8}. Historically, most research has focused on MSA; however, there has been a growing interest in DA within the research community in recent years, but research on the CA remains very sparse. 

During the Multi-Genre Broadcast (MGB2) Challenge \cite{article:mgb2}, the most well-known benchmark dataset for Arabic, MGB2, was released. The best-performing model in the challenge combined three 
acoustic models: Time-Delay Neural Network (TDNN)\cite{article:tdnn}, Long Short-Term Memory (LSTM) \cite{article:lstm}, and Bidirectional LSTM (BLSTM).
For language modeling, a tri-gram was used for the first-pass decoding, followed by an interpolated four-gram and a Recurrent Neural Network (RNN) with Maximum Entropy (MaxEnt) connections for rescoring. Model achieved a Word Error Rate (WER) of $14.7\%$. In \cite{article:e2e1}, Ahmet et al. introduced the first end-to-end recipe for Arabic ASR systems. Their system consists of a Bidirectional Recurrent Neural Network (BDRNN) trained with Connectionist Temporal Classification (CTC)\cite{article:ctc} objective function and a 16-gram LM, achieving a $12.03\%$ WER on the MGB2. In \cite{article:toolkit_espnet_recipe}, a transformer-based Arabic ASR system was introduced, which achieved a WER $12.5\%$ on the MGB2 dataset. In the recent work \cite{article:e2e7}, an Arabic ASR system utilizing self-supervised speech representations was introduced. They began by pretraining speech representations using Data2Vec \cite{article:data2vec} on the Aswat and MGB2 datasets, then finetuned the model for ASR using the MGB2 and Mozilla Common Voice (MCV) Arabic\cite{dataset:mcv} datasets. Reported WERs are $11.7\%$ and $10.3\%$ on MCV and the MGB2, respectively. In \cite{article:FLEURS} Talafha et al. performed N-shot finetuning of the Whisper model \cite{article:whisper} and reported $15.49\%$ on MGB2 and $10.36\%$ WER on FLEURS Arabic subset\cite{dataset:fleurs}. More recently, Toyin et al.\cite{article:artst} released a SpeechT5 model pre-trained on explicitly Arabic data. They fine-tuned it for various tasks, including ASR, and reported a WER of $12.78\%$ on the MGB2 test set.

With the growing interest in Arabic ASR, new benchmark datasets are being introduced. In recent work, Kolobov et al. presented the MediaSpeech dataset \cite{article:mediaspeech} reporting a WER of $13\%$ with Quartznet \cite{article:quartznet} model trained on a proprietary dataset. More recently, the Massive Arabic Speech Corpus (MASC) was released \cite{article:masc}, along with a pretrained DeepSpeech\cite{article:deepspeech} model, which achieved a WER of $21.8\%$ on the MASC test set.

In our work, we aim to advance further by not only providing accurate speech transcriptions but also enhancing model output readability through the inclusion of punctuation and diacritical marks. Additionally, we aim to provide a baseline for CA ASR.
\subsection{Open-source recipes}
Despite the growing interest in Arabic ASR, there have been relatively few comprehensive efforts to develop a full pipeline specifically for Arabic. One of the first Arabic ASR models was introduced by Ali et al.  \cite{article:toolkit_kaldi_recipe} using the Kaldi toolkit \cite{toolkit:kaldi}. Their system had three main parts: a DNN-based acoustic model, LM, and a lexicon. Training scripts, along with a data processing pipeline, were released.
In contrast, Hussein et al. \cite{article:toolkit_espnet_recipe} introduced a pipeline for end-to-end Arabic ASR using ESPnet toolkit \cite{toolkit:espnet}. Authors develop a transformer-based \cite{article:transformer} ASR model with a CTC / attention objective function. However, these models were trained on data that restricts commercial use, and the models themselves were not published.

Our models leverage a modified version of the Conformer architecture \cite{article:conformer} known as FastConformer \cite{article:fastconformer}. Conformers integrate convolutional layers with Transformers, and have been shown to outperform previous approaches that relied solely on Transformers. 




\section{Datasets and Data Processing}
\subsection{Datasets}
Our models were trained on four open-source datasets: MASC, MCV17.0 Arabic, the Arabic section of the FLEURS, and the Tarteel AI's EveryAyah dataset\cite{dataset:everyayah}. The primary dataset, MASC, is relatively new, with limited benchmarks available in the literature. To ensure comparability with other models, we report our model evaluation results on the MGB2 test set without using its training set (blind evaluation). Additionally, another publicly available dataset -- MediaSpeech was used for blind evaluation.

MASC dataset includes more than 1,000 hours of Arabic speech from over 700 YouTube channels. It is divided into clean and noisy sections. The development and test provided by dataset authors consist of about 10 hours of manually aligned and transcribed data, with diacritics and punctuation removed and letter normalization applied. Specifically, the variation of the letter Alif ``\<\smallأ إ آ ا>\,'' was normalized to a plain Alif ``\<\smallا>\,'', and the Ta' Marbuta ``\<\smallة>\,'' was normalized to Ha ``\<\smallه>\,''. Although this normalization approach reduces the impact of spelling inconsistency and misplaced letters, it increases text ambiguity and impacts readability. To evaluate model performance on the MASC dataset, we extracted development and test sets from the raw data based on the video IDs provided by the authors. In the tables below, the results on our extracted sets are labeled as ``MASC''. EveryAyah dataset is a collection of Quranic verses and their transcriptions, with diacritization. It contains high-quality recordings by professional reciters. This dataset was used only for CA model training. General information on datasets can be found in table \ref{tab:datasets}.

\begin{table}[ht]
    \caption{Dataset Descriptions.}
        \setlength{\tabcolsep}{1.8pt}
        \begin{center}
            \scriptsize
            \begin{tabularx}{\linewidth}{Xcccccc}
                \toprule
                 & \textbf{MASC} & \multirow{2}{*}{\textbf{MCV}} & \multirow{2}{*}{\textbf{FLEURS}} & \textbf{Every-} & \textbf{Media-} & \multirow{2}{*}{\textbf{MGB2}} \\
                 & \textbf{clean+noisy} & & & \textbf{Ayah} & \textbf{Speech} & \\
                \midrule
                \, \textbf{Raw Train (h)} & 646+855 & 90 & 6 & 829 & - & - \\
                \, \textbf{Raw Test (h)} & 13 & 12 & 1 & 104 & 13 & 10 \\
                \midrule 
                \, \textbf{Processed Train (h)} & 433+255 & 66 & 5 & 392 & - & - \\
                \, \textbf{Processed Test (h)} & 9 & 10 & 1 & 104 / 23 & 10 & 10 \\
                \midrule
                \, \textbf{Diacritization}  & Mix & Mix & Partial & Full & No & Partial \\
                \midrule
                \, \textbf{MSA}  & \tikzcmark & \tikzcmark & \tikzxmark    & \tikzxmark & \tikzcmark & \tikzcmark\\
                \, \textbf{CA}   & \tikzcmark & \tikzcmark & \tikzxmark    & \tikzcmark & \tikzxmark & \tikzxmark\\
                \, \textbf{DA}   & \tikzcmark & \tikzxmark & \tikzcmark & \tikzxmark & \tikzxmark & \tikzcmark \\
                \midrule
                \, \textbf{Punctuation} & \tikzcmark & \tikzcmark & \tikzcmark   & \tikzxmark & \tikzxmark & \tikzcmark \\
                \midrule
                \, \textbf{Letter norm.} & \tikzxmark & \tikzxmark & \tikzxmark & \tikzxmark & \tikzcmark & \tikzxmark \\
                \midrule
                \, \textbf{License} & CC-BY-4.0 & CC-0 & CC-BY-4.0 & MIT & CC-BY-4.0 & MGB2 \\
                \bottomrule
            \end{tabularx}
            \label{tab:datasets}
        \end{center}
\end{table}

\subsection{Data processing}
In addition to the Arabic alphabet's 28 primary letters, we included six forms of Hamza ''\,\<\small أ إ ء ؤ ئ آ >\,``, Ta' Marbuta ''\,\<\small ة>\,`` and Alif Maksura ''\,\<\small ى>\,``. Although these symbols are not part of the standard alphabet, their inclusion enhances readability. In total, our alphabet included 36 Arabic characters. Along with letters, we also considered the use of diacritics: Fathah, Kasrah, Dammah, Sukun, Tanween, and Shaddah. The alphabet with diacritics consists of 44 symbols. Moreover, the following punctuation marks were included: final stop ``.'', Arabic question mark ``\<\small؟>'', and Arabic comma ``\<\small،>''.

Our data preprocessing was done using NeMo-speech-data-processor \cite{toolkit:sdp}. Preprocessing steps are show in the table \ref{tab:preprocessing_steps}.

\begin{table}[ht]
    \centering
    \caption{Summary of preprocessing steps applied to the dataset}
    \begin{tabularx}{\linewidth}{X}
    \toprule
    \textbf{Basic:} \\
        ~ 1.~ Drop samples with out-of-alphabet symbols \\
        ~ 2.~ Drop samples shorter than 0.1s or longer than 20s \\
    \textbf{Data normalization:} \\
        ~ 1.~ Convert Eastern Arabic numerals to standard numerals \\
        ~ 2.~ Apply text normalization (NeMo-Text-Processing\cite{toolkit:nemo-text-processing}) \\
        ~ 3.~ Decompose ligatures into individual letters \\
        ~ 4.~ Apply NFKC normalization \\
        ~ 5.~ Remove rare punctuation marks (ex: ``:'', ``-'', quotation marks)  \\
        ~ 6.~ Remove white-spaces before punctuation marks \\
        ~ 7.~ Replace punctuation with Arabic marks and remove repetitions \\
        ~ 8.~ Remove Kasheeda mark ``\<ـ>'' \\
    \textbf{Data filtration:} \\
        ~ 1.~ Drop samples with high and low word rates \\
        ~ 2.~ Drop samples with high and low char rates \\
        ~ 3.~ Use pretrained model and drop high WER and CER samples \\
    \bottomrule
    \end{tabularx}
    \label{tab:preprocessing_steps}
\end{table}

Data normalization is done to standardize the model's alphabet to preserve more relevant data. Arabic letters can appear in different forms depending on their position within a word. NFKC normalization is done to replace these positional forms with their corresponding general Unicode representations. Additionally, this normalization form ensures a consistent ordering of diacritical marks in words where multiple diacritics may accompany a single letter. Data filtration steps are done to remove samples with noise, music, or non-Arabic speech. The effectiveness of chosen approach is demonstrated in the table \ref{tab:wer_reduction}. For the MASC dataset we were able to extract additional 120h data from clean subset and 40h from noisy. The last row shows the final setup, where the noisy subset of MASC was further filtered. More details are provided in the Results section.

Before this step, MASC data was split into segments by timestamps contained in .vtt files. These segments were then combined into samples, each with a maximum duration of 20 seconds. Additionally, numeration was removed from samples where it was present in the text.

The MCV17 dataset had overlapping samples between train, test, and dev sets, where the same sentences appeared with different punctuation and diacritics. These overlaps were removed from the training set but retained in the dev and test sets. During the MediaSpeech dataset processing, we found it contained only 31 unique Arabic symbols. In addition to discussed letter normalization, Alif Maksura ``\<\smallى>''  was replaced by Ya' ``\<\smallي>''. We applied similar letter normalization for consistency in evaluation. For the EveryAyah dataset, overlap between train and test sets was detected, where different speakers recorded identical transcriptions. To resolve overlaps, we removed diacritics and punctuation, then excluded any test set texts also found in the training set. The test set was reduced from $100$h to $23$h (see table \ref{tab:datasets}). In the sections below results for both sets are shown.


\section{Experimental Setup}

In this study, we utilized the FastConformer hybrid RNN-T (Recurrent Neural Network Transducer \cite{article:rnnt}) with the CTC model.
We set the dropout rate to 0.1 to avoid overfitting and augmented the training data using spectral augmentation \cite{article:specaug} with 4 frequency and 20 time masks. The model combines two decoders, RNN-T and CTC, with the CTC weight set to $0.3$. Mean batch reduction is applied to both decoders. We use a SentencePiece tokenizer with a vocabulary of 1,024 tokens. For model training, the AdamW optimizer with CosineAnnealing scheduler was used. Best results were obtained with learning rate $\mathrm{lr}=5\cdot10^{-3}$ and $2000$ warmup steps. Final models were trained for $200$ epochs. The weights of the five best checkpoints were averaged.

In addition to traditional WER metric, we also considered two other WER metrics\cite{article:mgb2}. In the following sections, we use WER\textsubscript{PC,D} -- scored with punctuated and diacritized text, WER\textsubscript{PC} -- scored with just punctuated text, and WER -- scored after removing punctuation and diacritics.

During training, we tested different starting checkpoints for the encoder. Initially, we trained the model from scratch with randomly initialized weights and a learning rate of $\mathrm{lr}=2\cdot10^{-3}$ to ensure convergence. We then used three models, English\cite{model:en}, Spanish\cite{model:es}, and Multilingual as starting checkpoints for the Arabic model, as they were trained on extensive datasets and, likely, capture linguistic features transferable across languages. As detailed in Table \ref{tab:starting_checkpoints}, the checkpoints produced similar results. The Spanish checkpoint was chosen for faster convergence, likely due to similarities in phonetics and syntax between Spanish and Arabic.

\begin{table}[h]
\caption{Model performance (WER \%) for different starting checkpoints.}
    \begin{center}
        \begin{tabularx}{\linewidth}{XYYY}
            \toprule
            \textbf{Starting} & \multicolumn{3}{c}{\textbf{Test set}} \\
            \textbf{Checkpoint} & \textbf{MASC}& \textbf{MCV}& \textbf{FLEURS} \\
            \midrule
            From scratch    & 12.74 & 17.53 & 12.38 \\
            English         & 12.23 & 10.67 & 8.37  \\
            Spanish         & \textbf{11.63}& \textbf{10.21} & 8.18 \\
            Multilingual    & 11.66 & 10.53 & \textbf{8.12} \\
            \bottomrule
        \end{tabularx}
        \label{tab:starting_checkpoints}
    \end{center}
\end{table}

\begin{table}[ht]
\caption{WERs (\%) for different data preprocessing. B-basic, O-Our}
\setlength{\tabcolsep}{3.pt}
    \begin{center}
        \begin{tabularx}{1.\linewidth}{Xccc}
            \toprule
            \textbf{Training} & \multicolumn{3}{c}{\textbf{Test set}} \\
            \textbf{Sets} & \textbf{MASC}& \textbf{MCV}& \textbf{FLEURS} \\
            \midrule
            B: MCV+FLEURS+MASC\textsubscript{cln(313h)} & 11.32 & 9.50 & 6.10 \\
            B: MCV+FLEURS+MASC\textsubscript{cln(313h)+nsy(353h)} & 11.55 & 8.97 & 6.40 \\
            \midrule
            O: MCV+FLEURS+MASC\textsubscript{cln(433h)} & 9.25 & 8.90 & 6.19 \\
            O: MCV+FLEURS+MASC\textsubscript{cln(433h)+nsy(395h)} & 9.95 & 8.68 & 5.76 \\
            O: MCV+FLEURS+MASC\textsubscript{cln(433h)+extr. nsy(255h)} & \textbf{8.50} & \textbf{8.20} & \textbf{5.01} \\
            \bottomrule
        \end{tabularx}
        \label{tab:wer_reduction}
    \end{center}
\end{table}

\begin{table*}[ht]
\caption{Comparing performance against Whisper, Seamless, ArTST and best performing models reported previously.
EA denotes EveryAyah. For the EveryAyah dataset, WER values are reported both before / after overlap removal. For MediaSpeech dataset letter normalization was applied in post-processing step.}
    \begin{center}
        \begin{tabularx}{0.85\linewidth}{l|Xcccc|cc}
            \toprule   
            & & \textbf{MASC} & \textbf{MCV} & \textbf{FLEURS} & \textbf{EveryAyah} & \textbf{MediaSpeech*} & {\textbf{MGB2}}  \\
            \midrule
            \parbox[t]{3mm}{\multirow{7}{*}{\rotatebox[origin=c]{90}{WER}}} & Reported SOTA & 21.8\cite{article:masc} & 11.7\cite{article:e2e7} & 10.30\cite{article:FLEURS} & - & \textbf{13.00} \cite{article:mediaspeech} & \textbf{10.36} \cite{article:e2e7}  \\
            & Whisper-Large-v3 & 13.82 & 16.74 & 8.28 & 9.37 / 17.17 & 22.00 & 21.52 \\
            & SeamlessM4T-Large-v2 & 15.50 & 12.22 & 6.32 & - / 11.86 & 18.10 & 18.53 \\
            & ArTST  & 23.58 & 36.16 & 15.53 & - / 58.82 & 22.34 & 12.78 \\
            & \textbf{Our}: no PC \& no Diacr & \textbf{8.50} & 8.20 & \textbf{5.01} & 6.28 / 8.47 & 21.90 & 23.4 \\
            & \textbf{Our}: PC \& no Diacr & 8.52 & \textbf{7.97}  & 5.08 & 6.40 / 6.53  & 21.63 & 21.11\\
            & \textbf{Our}: PC \& Diacr (w/o EA) & 8.59 & 8.66  & 5.07 & 6.14 / 7.75  & 22.61 & 23.97 \\
            & \textbf{Our}: PC \& Diacr & 8.64 & 8.52  & 5.39 & 1.34 / \textbf{5.73}  & 22.75 & 24.70 \\
            \midrule
            \parbox[t]{3mm}{\multirow{3}{*}{\rotatebox[origin=c]{90}{WER\textsubscript{PC}}}} & Whisper-Large-v3 & 27.79 & 41.17 & 22.36 & 9.38 / 17.18 & - & 21.79 \\
            & SeamlessM4T-Large-v2 & 29.77 & 35.04 & 15.52 & - / 11.90 & - & 21.37 \\
            & \textbf{\underline{Our MSA}}: PC \& no Diacr & 11.63 & 10.21  & 8.18 & 6.40 / 6.53  & - & 21.17 \\
            \midrule
            \parbox[t]{3mm}{\multirow{4}{*}{\rotatebox[origin=c]{90}{WER\textsubscript{PC,D}}}} & Whisper-Large-v3 & 13.82 & 16.74 & 8.28 & 98.78 / 99.40 & - & 21.90 \\
            & SeamlessM4T-Large-v2 & 15.50 & 12.22 & 6.32 & - / 98.42 & - & 22.45 \\
            & \textbf{Our}: PC \& Diacr (w/o EA) & 16.51 & 25.41  & 13.09 & 91.60 / 90.49  & - & 23.97 \\
            & \textbf{\underline{Our CA}}: PC \& Diacr & 16.67 & 25.60  & 12.94 & 1.55 / 6.65  & - & 25.58\\
            \bottomrule
        \end{tabularx}
        \label{tab:wers}
    \end{center}
\end{table*}

\section{Results}
\subsection{Extracting clean data from MASC noisy subset}
To extract additonal clean data from MASC noisy subset, we begun by training a model on the clean training set of the MASC, MCV, and FLEURS datasets, after removing punctuation marks and diacritics. This model was then used to perform inference on the noisy MASC data. Samples with a WER greater than 60 or a Character Error Rate (CER) above 30 were discarded. Through this process, an additional 255 hours of clean data were extracted and added to the training sets, leading to a reduction in WER as shown in the final row of Table \ref{tab:wer_reduction}.

\subsection{Models}
The first model \textit{Our MSA: PC \& no Diacr}, trained on the MASC, MCV, and FLEURS datasets, is designed for MSA speech recognition. Since MSA text typically lacks diacritical marks, the model outputs punctuated but undiacritized text. To examine the impact of including punctuation marks in overall model accuracy, a baseline model without punctuation and diacritical marks: \textit{Our: no PC \& no Diacr} was trained. Evaluation results are shown in Table \ref{tab:wers}. Baseline and punctuated models achieve similar $\mathrm{WER}$ scores, indicating that while adding punctuation increases the WER, it does not negatively affect word recognition quality. Furthermore, our model outperforms the best systems reported in the literature on in-domain datasets and delivers comparable results on out-of-domain data. In all setups, our model still outperforms larger counterparts like Whisper-Large-v3 and SeamlessM4T-Large-v2 \cite{article:seamless}, that were trained on larger, proprietary datasets.

Our second model \textit{Our CA: PC \& Diacr}, trained on the MASC, MCV, FLEURS, and EveryAyah datasets, is designed for both MSA and CA speech recognition and supports diacritical and punctuation marks. The evaluation results are presented in Table \ref{tab:wers}. Including diacritics increased the overall WER\textsubscript{PC,D}, primarily because diacritical marks in MSA serve as supplementary phonetic guides and lack standardized rules. Moreover, we observed inconsistencies in diacritization across the training and evaluation datasets. This issue is particularly evident on the MCV17 dataset, where some samples in the dev and test sets contain undiacritized texts, while others include full diacritical marks. However, a comparison with the WER values of our baseline model suggests that including diacritics in the training data has a minor impact on overall recognition quality.

Our baseline model, trained on the MASC, MCV, and FLEURS datasets, shows acceptable results for CA speech recognition, achieving a WER of approximately $6.14\%$ on the EveryAyah test set. However, it exhibits a very high WER\textsubscript{PC,D} on the same set due to its tendency to produce partially diacritized text. Adding the EveryAyah dataset to the training set improved the model's ability to recognize CA speech, which led to $\mathrm{WER}\textsubscript{PC,D}$ drop to $1.55\% / 6.65\%$ on the EveryAyah test set, with only minor changes to this metric for other test sets.

\section{Conclusion}
In this paper, we presented the development of two Arabic ASR models using open-source datasets, along with a data processing pipeline and a detailed analysis of available datasets. The first model, designed for MSA, is the first end-to-end unified Arabic ASR and punctuation model to be released as open-source. The model achieves competitive WERs: $11.37\%$ on MASC, $9.76\%$ on MCV Arabic, and $7.73\%$ on FLEURS Arabic subset. The second model, a unified MSA and CA system, achieves a diacritic- and punctuation-aware WER of $6.65\%$ on the EveryAyah dataset, while still delivering strong performance for MSA: $16.67\%$ on MASC, $25.60\%$ on MCV Arabic and $12.94\%$ on FLEURS dataset.

\newpage
\clearpage
\bibliographystyle{IEEEtran}
\bibliography{mybib.bib}

\begin{thebibliography}{10}
\providecommand{\url}[1]{#1}
\csname url@samestyle\endcsname
\providecommand{\newblock}{\relax}
\providecommand{\bibinfo}[2]{#2}
\providecommand{\BIBentrySTDinterwordspacing}{\spaceskip=0pt\relax}
\providecommand{\BIBentryALTinterwordstretchfactor}{4}
\providecommand{\BIBentryALTinterwordspacing}{\spaceskip=\fontdimen2\font plus
\BIBentryALTinterwordstretchfactor\fontdimen3\font minus \fontdimen4\font\relax}
\providecommand{\BIBforeignlanguage}[2]{{%
\expandafter\ifx\csname l@#1\endcsname\relax
\typeout{** WARNING: IEEEtran.bst: No hyphenation pattern has been}%
\typeout{** loaded for the language `#1'. Using the pattern for}%
\typeout{** the default language instead.}%
\else
\language=\csname l@#1\endcsname
\fi
#2}}
\providecommand{\BIBdecl}{\relax}
\BIBdecl

\bibitem{toolkit:nemo}
\BIBentryALTinterwordspacing
N.~Corporation, ``Nvidia nemo: A toolkit for conversational ai,'' 2020. [Online]. Available: \url{https://developer.nvidia.com/nvidia-nemo}
\BIBentrySTDinterwordspacing

\bibitem{article:hmm1}
M.~Nofal, E.~Reheem, H.~Henawy, and N.~Kader, ``The development of acoustic models for command and control arabic speech recognition system,'' 01 2004.

\bibitem{article:hmm2}
F.~Elmisery, A.~Khalil, A.~Salama, and H.~Hammed, ``A fpga-based hmm for a discrete arabic speech recognition system,'' pp. 322--325, 2003.

\bibitem{article:hmm3}
B.~Hocine, R.~Djemili, M.~Bedda, and C.~Snani, ``New hybrid system (supervised classifier/hmm) for isolated arabic speech recognition,'' 01 2006, pp. 1264 -- 1269.

\bibitem{article:toolkit_kaldi_recipe}
A.~Ali, Y.~Zhang, P.~Cardinal, N.~Dahak, S.~Vogel, and J.~Glass, ``A complete kaldi recipe for building arabic speech recognition systems,'' \emph{2014 IEEE Workshop on Spoken Language Technology, SLT 2014 - Proceedings}, pp. 525--529, 04 2015.

\bibitem{article:gmm1}
A.~E. Kourd and K.~E. Kourd, ``Arabic isolated word speaker dependent recognition system,'' \emph{Journal of Advances in Mathematics and Computer Science}, vol.~14, no.~1, p. 1–15, Jan. 2016.

\bibitem{article:dnn1}
P.~Cardinal, A.~Ali, N.~Dehak, Y.~Zhang, T.~Hanai, J.~Glass, and S.~Vogel, ``Recent advances in asr applied to an arabic transcription system for al-jazeera,'' \emph{Proceedings of the Annual Conference of the International Speech Communication Association, INTERSPEECH}, pp. 2088--2092, 01 2014.

\bibitem{article:dnn2}
T.~AlHanai, W.-N. Hsu, and J.~Glass, ``Development of the mit asr system for the 2016 arabic multi-genre broadcast challenge,'' in \emph{2016 IEEE Spoken Language Technology Workshop (SLT)}, 2016, pp. 299--304.

\bibitem{article:hybrid1}
A.~Ouisaadane and S.~Said, ``A comparative study for arabic speech recognition system in noisy environments,'' \emph{International Journal of Speech Technology}, vol.~24, pp. 1--10, 09 2021.

\bibitem{article:hybrid2}
A.~Masmoudi, F.~Bougares, M.~Ellouze, Y.~Estève, and L.~Belguith, ``Automatic speech recognition system for tunisian dialect,'' \emph{Language Resources and Evaluation}, vol.~52, 03 2018.

\bibitem{article:hybrid3}
M.~Elmahdy, M.~Hasegawa-Johnson, and E.~Mustafawi, ``Development of a {TV} broadcasts speech recognition system for qatari {A}rabic,'' in \emph{Proceedings of the Ninth International Conference on Language Resources and Evaluation ({LREC}'14)}, 2014, pp. 3057--3061.

\bibitem{article:e2e1}
A.~Ahmed, Y.~Hifny, K.~Shaalan, and S.~Toral, \emph{End-to-End Lexicon Free Arabic Speech Recognition Using Recurrent Neural Networks}, 11 2018, pp. 231--248.

\bibitem{article:e2e3}
N.~Zerari, S.~Abdelhamid, H.~Bouzgou, and C.~Raymond, ``Bidirectional deep architecture for arabic speech recognition,'' \emph{Open Computer Science}, vol.~9, no.~1, pp. 92--102, 2019.

\bibitem{article:e2e4}
Y.~Belinkov, A.~Ali, and J.~Glass, ``Analyzing phonetic and graphemic representations in end-to-end automatic speech recognition,'' 2020.

\bibitem{article:e2e5}
H.~A. Alsayadi, A.~A. Abdelhamid, I.~Hegazy, and Z.~T. Fayed, ``Arabic speech recognition using end-to-end deep learning,'' \emph{IET Signal Process.}, vol.~15, pp. 521--534, 2021.

\bibitem{article:e2e6}
S.~A. Chowdhury, A.~Hussein, A.~Abdelali, and A.~Ali, ``{Towards One Model to Rule All: Multilingual Strategy for Dialectal Code-Switching Arabic ASR},'' in \emph{Proc. Interspeech 2021}, 2021, pp. 2466--2470.

\bibitem{article:e2e7}
L.~Alkanhal, A.~Alessa, E.~Almahmoud, and R.~Alaqil, ``Aswat: {A}rabic audio dataset for automatic speech recognition using speech-representation learning,'' in \emph{Proceedings of ArabicNLP 2023}, 2023, pp. 120--127.

\bibitem{article:e2e8}
A.~Hussein, S.~Watanabe, and A.~Ali, ``Arabic speech recognition by end-to-end, modular systems and human,'' 2021.

\bibitem{article:mgb2}
A.~Ali, P.~Bell, J.~Glass, Y.~Messaoui, H.~Mubarak, S.~Renals, and Y.~Zhang, ``The mgb-2 challenge: Arabic multi-dialect broadcast media recognition,'' 12 2016, pp. 279--284.

\bibitem{article:tdnn}
V.~Peddinti, D.~Povey, and S.~Khudanpur, ``A time delay neural network architecture for efficient modeling of long temporal contexts,'' in \emph{Interspeech 2015}, 2015, pp. 3214--3218.

\bibitem{article:lstm}
H.~Sak, A.~Senior, and F.~Beaufays, ``Long short-term memory based recurrent neural network architectures for large vocabulary speech recognition,'' 2014.

\bibitem{article:ctc}
A.~Graves, S.~Fern\'{a}ndez, F.~Gomez, and J.~Schmidhuber, ``Connectionist temporal classification: labelling unsegmented sequence data with recurrent neural networks,'' in \emph{Proceedings of the 23rd International Conference on Machine Learning}, ser. ICML '06.\hskip 1em plus 0.5em minus 0.4em\relax New York, NY, USA: Association for Computing Machinery, 2006, p. 369–376.

\bibitem{article:toolkit_espnet_recipe}
A.~Hussein, S.~Watanabe, and A.~Ali, ``Arabic speech recognition by end-to-end, modular systems and human,'' \emph{Computer Speech \& Language}, vol.~71, p. 101272, 2022.

\bibitem{article:data2vec}
A.~Baevski, W.-N. Hsu, Q.~Xu, A.~Babu, J.~Gu, and M.~Auli, ``data2vec: A general framework for self-supervised learning in speech, vision and language,'' 2022.

\bibitem{dataset:mcv}
R.~Ardila, M.~Branson, K.~Davis \emph{et~al.}, ``Common voice: A massively-multilingual speech corpus,'' in \emph{Proceedings of the Twelfth Language Resources and Evaluation Conference}, 2020, pp. 4218--4222.

\bibitem{article:FLEURS}
B.~Talafha, A.~Waheed, and M.~Abdul-Mageed, ``{N-Shot Benchmarking of Whisper on Diverse Arabic Speech Recognition},'' in \emph{Proc. INTERSPEECH 2023}, 2023, pp. 5092--5096.

\bibitem{article:whisper}
A.~Radford, J.~W. Kim, T.~Xu, G.~Brockman, C.~McLeavey, and I.~Sutskever, ``Robust speech recognition via large-scale weak supervision,'' 2022.

\bibitem{dataset:fleurs}
A.~Conneau, M.~Ma, S.~Khanuja, Y.~Zhang, V.~Axelrod, S.~Dalmia, J.~Riesa, C.~Rivera, and A.~Bapna, ``Fleurs: Few-shot learning evaluation of universal representations of speech,'' \emph{2022 IEEE Spoken Language Technology Workshop (SLT)}, pp. 798--805, 2022.

\bibitem{article:artst}
H.~Toyin, A.~Djanibekov, A.~Kulkarni, and H.~Aldarmaki, ``{A}r{TST}: {A}rabic text and speech transformer,'' in \emph{Proceedings of ArabicNLP 2023}.\hskip 1em plus 0.5em minus 0.4em\relax Singapore (Hybrid): Association for Computational Linguistics, Dec. 2023, pp. 41--51.

\bibitem{article:mediaspeech}
R.~Kolobov, O.~Okhapkina, A.~P. Olga~Omelchishina, R.~Bedyakin, V.~Moshkin, D.~Menshikov, and N.~Mikhaylovskiy, ``Mediaspeech: Multilanguage asr benchmark and dataset,'' 2021.

\bibitem{article:quartznet}
S.~Kriman, S.~Beliaev, B.~Ginsburg, J.~Huang, O.~Kuchaiev, V.~Lavrukhin, R.~Leary, J.~Li, and Y.~Zhang, ``Quartznet: Deep automatic speech recognition with 1d time-channel separable convolutions,'' 2019.

\bibitem{article:masc}
M.~Al-Fetyani, M.~Al-Barham, G.~Abandah, A.~Alsharkawi, and M.~Dawas, ``Masc: Massive arabic speech corpus,'' in \emph{2022 IEEE Spoken Language Technology Workshop (SLT)}, 2023, pp. 1006--1013.

\bibitem{article:deepspeech}
A.~Hannun, C.~Case, J.~Casper, B.~Catanzaro, G.~Diamos, E.~Elsen, R.~Prenger, S.~Satheesh, S.~Sengupta, A.~Coates, and A.~Y. Ng, ``Deep speech: Scaling up end-to-end speech recognition,'' 2014.

\bibitem{toolkit:kaldi}
\BIBentryALTinterwordspacing
D.~Povey, A.~Ghoshal, G.~Boulianne, L.~Zettlemoyer, V.~Panayotov \emph{et~al.}, ``Kaldi speech recognition toolkit,'' 2011. [Online]. Available: \url{http://kaldi-asr.org/}
\BIBentrySTDinterwordspacing

\bibitem{toolkit:espnet}
S.~Watanabe, T.~Hori, S.~Karita, T.~Hayashi, J.~Nishitoba, Y.~Unno, N.~E.~Y. Soplin, J.~Heymann, M.~Wiesner, N.~Chen, A.~Renduchintala, and T.~Ochiai, ``Espnet: End-to-end speech processing toolkit,'' 2018.

\bibitem{article:transformer}
A.~Vaswani, N.~Shazeer, N.~Parmar, J.~Uszkoreit, L.~Jones, A.~N. Gomez, L.~Kaiser, and I.~Polosukhin, ``Attention is all you need,'' in \emph{Proceedings of the 31st International Conference on Neural Information Processing Systems}, ser. NIPS'17.\hskip 1em plus 0.5em minus 0.4em\relax Curran Associates Inc., 2017, p. 6000–6010.

\bibitem{article:conformer}
A.~Gulati, J.~Qin, C.-C. Chiu, N.~Parmar, Y.~Zhang, J.~Yu, W.~Han, S.~Wang, Z.~Zhang, Y.~Wu, and R.~Pang, ``Conformer: Convolution-augmented transformer for speech recognition,'' 2020.

\bibitem{article:fastconformer}
D.~Rekesh, N.~R. Koluguri, S.~Kriman, S.~Majumdar, V.~Noroozi, H.~Huang, O.~Hrinchuk, K.~Puvvada, A.~Kumar, J.~Balam, and B.~Ginsburg, ``Fast conformer with linearly scalable attention for efficient speech recognition,'' 2023.

\bibitem{dataset:everyayah}
``Tarteel ai's everyayah dataset,'' \url{https://hf.co/datasets/tarteel-ai/everyayah}, accessed: 2024-09-12.

\bibitem{toolkit:sdp}
\BIBentryALTinterwordspacing
N.~Corporation, ``Speech data processor (sdp) toolkit,'' 2024. [Online]. Available: \url{https://github.com/NVIDIA/NeMo-speech-data-processor}
\BIBentrySTDinterwordspacing

\bibitem{toolkit:nemo-text-processing}
Y.~Zhang, E.~Bakhturina, and B.~Ginsburg, ``{NeMo (Inverse) Text Normalization: From Development to Production},'' in \emph{Proc. Interspeech 2021}, 2021, pp. 4857--4859.

\bibitem{article:rnnt}
A.~Graves, ``Sequence transduction with recurrent neural networks,'' 2012.

\bibitem{article:specaug}
D.~S. Park, W.~Chan, Y.~Zhang, C.-C. Chiu, B.~Zoph, E.~D. Cubuk, and Q.~V. Le, ``Specaugment: A simple data augmentation method for automatic speech recognition,'' in \emph{Interspeech 2019}.\hskip 1em plus 0.5em minus 0.4em\relax ISCA, Sep. 2019.

\bibitem{model:en}
\BIBentryALTinterwordspacing
N.~Corporation, ``Stt en fastconformer hybrid transducer-ctc large p\&c,'' 2020. [Online]. Available: \url{https://hf.co/nvidia/stt_en_fastconformer_hybrid_large_pc}
\BIBentrySTDinterwordspacing

\bibitem{model:es}
\BIBentryALTinterwordspacing
------, ``Stt es fastconformer hybrid transducer-ctc large p\&c,'' 2020. [Online]. Available: \url{https://hf.co/nvidia/stt_es_fastconformer_hybrid_large_pc}
\BIBentrySTDinterwordspacing

\bibitem{article:seamless}
``Seamlessm4t: Massively multilingual \& multimodal machine translation,'' 2023.

\end{thebibliography}
\end{document}